\documentclass[letterpaper,10pt,conference]{ieeeconf}

\IEEEoverridecommandlockouts
\usepackage{times}
\usepackage{mathptmx}
\usepackage{amsmath}
\usepackage{amssymb}
\usepackage{mathtools}
\usepackage{graphicx}
\usepackage{xcolor}
\usepackage{booktabs}
\usepackage{multirow}
\usepackage{tabularx}
\usepackage{array}
\usepackage[noadjust]{cite}

\let\olditemize\itemize
\let\endolditemize\enditemize
\renewenvironment{itemize}
{\olditemize\setlength{\itemsep}{0pt}\setlength{\parskip}{0pt}\setlength{\parsep}{0pt}}
{\endolditemize}

\title{\LARGE \bf
Reinforcement Learning with Inner-loop Dynamics Estimator for Aerial Manipulation under Uncertainty
}

\author{
Shivansh P. Singh$^{1*}$, 
Samaksh Ujjwal$^{1*}$, 
Ishita Chaudhary$^{1}$, 
V. R. Vasudevan$^{1}$, 
Rishabh Yadav$^{2}$, 
Spandan Roy$^{1}$%
\thanks{This work is partly supported by ``Edge-AI-GGCNN'' project sponsored by Qualcomm Technologies. $^{*}$ denotes equal contribution.}%
\thanks{$^{1}$International Institute of Information Technology Hyderabad, India. 
Emails: {\footnotesize\ttfamily shivansh.singh@research.iiit.ac.in},
{\footnotesize\ttfamily samaksh.ujjawal@research.iiit.ac.in},
{\footnotesize\ttfamily vasudevanvr2002@gmail.com}; 
{\footnotesize\ttfamily ishitachaudhary2003@gmail.com},
{\footnotesize\ttfamily spandan.roy@iiit.ac.in}.}%
\thanks{$^{2}$Department of Computer Science, University of Manchester, UK. 
Email: {\footnotesize\ttfamily rishabh.yadav@postgrad.manchester.ac.uk}.}%
}

\begin{document}

\maketitle
\thispagestyle{empty}
\pagestyle{empty}

\begin{abstract}
Aerial manipulators enable physical interaction in hard-to-reach environments; however, the combined problem of direct whole-body aerial manipulation under rapid arm motion, payload changes, and related unknown dynamic uncertainty remains a largely unsolved problem.
We present a hierarchical control framework that combines Reinforcement Learning (RL) with an inner-loop dynamics estimator to address this problem. The RL outer loop maps desired 6-degrees-of-freedom (DOF) end-effector targets to coordinated whole-body commands, enabling direct task-driven control without relying on a fully accurate coupled dynamic model in the policy layer. An inner loop then tracks these commands while compensating for transient inertial shifts and uncertainty during execution via a dynamics estimator scheme without requiring system model knowledge. We validate the proposed approach on a custom quadrotor equipped with a 3-DoF manipulator through hardware experiments under varying payload conditions. Compared with RL+PID and RL+INDI+PID baselines, the proposed method reduces end-effector tracking error and improves task success rate across the tested hardware conditions. These results show that combining learned whole-body coordination with estimator-based low-level compensation improves the precision and robustness of aerial manipulation under changing operating conditions.
\end{abstract}

\section{Introduction}
Aerial manipulators integrate a multi-rotor UAV with a robotic arm, enabling physical interaction in complex 3D environments, including inspection and contact-rich pick-and-place in confined or hard-to-reach spaces~\cite{ollero2022past,orsag2017dexterous}. Reliable deployment remains challenging because arm motion strongly couples with the underactuated aerial base, causing shifts in center of gravity, inertia, and disturbances during grasping, release, or rapid motion~\cite{sharma2025impedance, yadav2025integrated,Yilmaz2019}. These uncertainties make stable and precise aerial manipulation difficult. Existing control methods are broadly categorized into model-based and learning-based approaches, reviewed below.

% Aerial manipulators, which combine a multi-rotor uncrewed aerial vehicle (UAV) with a robotic arm, extend robotic operations beyond passive observation to physical interaction in complex three-dimensional environments. This capability is important for inspection and contact-rich pick-and-place tasks in confined or difficult-to-reach environments~\cite{ollero2022past,orsag2017dexterous}. Yet, despite their practical promise, reliable deployment remains limited by a fundamental control difficulty: the system must maintain stable flight while executing dynamic manipulation with a moving arm.

% The main control challenge arises from the strong coupling between the underactuated aerial base and the manipulator~\cite{lippiello2012exploiting}. As the arm moves, the coupled system undergoes continuous changes in center of gravity and effective inertia, and these changes become more severe during grasping, release, rapid arm motion, or arm-induced disturbances~\cite{pounds2011grasping,Yilmaz2019}. The resulting combination of parametric uncertainty and unmodeled disturbance makes precise and stable aerial manipulation difficult in practice. Control of aerial manipulators lies at the intersection of underactuated flight, robotic manipulation, and uncertainty-aware control. Prior work can be broadly organized into two directions: model-based control and learning based control. Related work in this area is discussed below.

\subsection{Related Works}

\subsubsection{Model-based Control for Aerial Manipulators}
Recent model-based methods address uncertainty and external disturbances, such as wind, using robust control approaches~\cite{8924899, 8047452, 9802632}, as well as adaptive methods~\cite{10466505, 10505853, 9768118,7395364, 9812607}. However, analytically modeling center-of-mass shifts, mass/inertia redistribution, and UAV--manipulator coupling remains extremely difficult~\cite{yadav2024modular}. These methods are also mainly designed for tracking, interaction regulation, or grasp stabilization, rather than direct whole-body action generation from manipulation objectives.

% Various model-based approaches have been recently developed to tackle model uncertainties and external disturbances (e.g., wind) either by robust control methods such as linear estimator \cite{8924899}, extended high-gain observer \cite{8047452}, disturbance observer \cite{9139470}, RISE-based method \cite{9802632}, or by adaptive control methods such as adaptive disturbance observer \cite{9565382, 10466505}, adaptive backstepping \cite{10505853}, adaptive sliding mode observer \cite{9768118}, adaptive sliding mode control \cite{7395364, 9812607}. Unfortunately, it is very difficult, if at all possible, to analytically model dynamic changes in the center of mass and mass/inertia distribution and the dynamic coupling forces between the UAV and the manipulator (cf. {\cite[Ch.5.3]{Orsag2018AerialMD}}). Moreover, such methods are primarily designed for reference tracking, interaction regulation, or grasp stabilization rather than direct whole-body action generation from a manipulation objective.

\subsubsection{Learning-based Control for Aerial Manipulation}
To reduce reliance on exact analytical models, recent studies have introduced learning the residual dynamic model~\cite{cao2024computation, ujjawal2026learn, yadav2026learning, ujjawal2025aermani, yadav2025arcade, yadav2026physics}. Recent works also use language and vision models for grasping, placement, and task-level reasoning, including clutter-aware aerial grasping, language-grounded placement, and VLM-based skill selection~\cite{singh2026aerograb,mishra2026aeroplace, mishra2025aermani}. 
Directly learning the control is promising for aerial manipulation because it can learn coordinated policies from interaction while reducing dependence on accurate analytical models~\cite{alzorgan2023actuator,Cuniato2023,deshmukh2025global,Liu2022,nieto2024safe}. RL is particularly useful for mapping end-effector objectives to coordinated UAV--manipulator commands. However, decoupled RL approaches treat the UAV and arm separately, ignoring unavoidable coupling and degrading performance~\cite{Liu2022,nieto2024safe,ollero2022past,deshmukh2025global}. Recent work addresses this by learning whole-body commands for both the aerial base and manipulator~\cite{deshmukh2025global}, but still relies on inner-loop controllers based on simplified linear models. This remains limiting because aerial manipulator dynamics are highly nonlinear, strongly coupled, and difficult to model under grasp/release inertia changes, rapid arm motion, and arm-induced disturbances~\cite[Ch.5.3]{orsag2017aerial}.

% Learning-based control has emerged as a promising direction for aerial manipulation because it can learn coordinated policies directly from interaction while reducing reliance on fully accurate analytical models~\cite{Alzorgan2023, Cuniato2023, deshmukh2025global, Liu2022, nieto2024safe}. In particular, Reinforcement Learning (RL) is attractive for mapping end-effector objectives to coordinated commands for the aerial base and manipulator. The works \cite{Liu2022, nieto2024safe} follow a decoupled approach considering the UAV and the arm as separate unconnected systems; such a convention ignores the inescapable coupled dynamics terms between the UAV and arm leading to compromised performance \cite{ollero2022past, deshmukh2025global}. Recently, \cite{deshmukh2025global} has demonstrated RL-based framework to map a desired end-effector objective to coordinated commands for both the aerial base and the manipulator. Such frameworks use a combination of RL and an inner-loop controller; here the inner-loop controller converts the RL outputs-UAV attitude, body rate and arm angles into deployable rotor input and arm joint torques assuming linear system model. However, it is well known that in addition to being highly coupled and nonlinear, aerial manipulator dynamics are difficult to model analytically due to abrupt grasp/release-induced inertial variation, rapid manipulator transients, or arm-induced disturbances during execution {\cite[Ch.5.3]{Orsag2018AerialMD}}.

\begin{figure*}[t]
    \centering
    \includegraphics[width=0.82\textwidth]{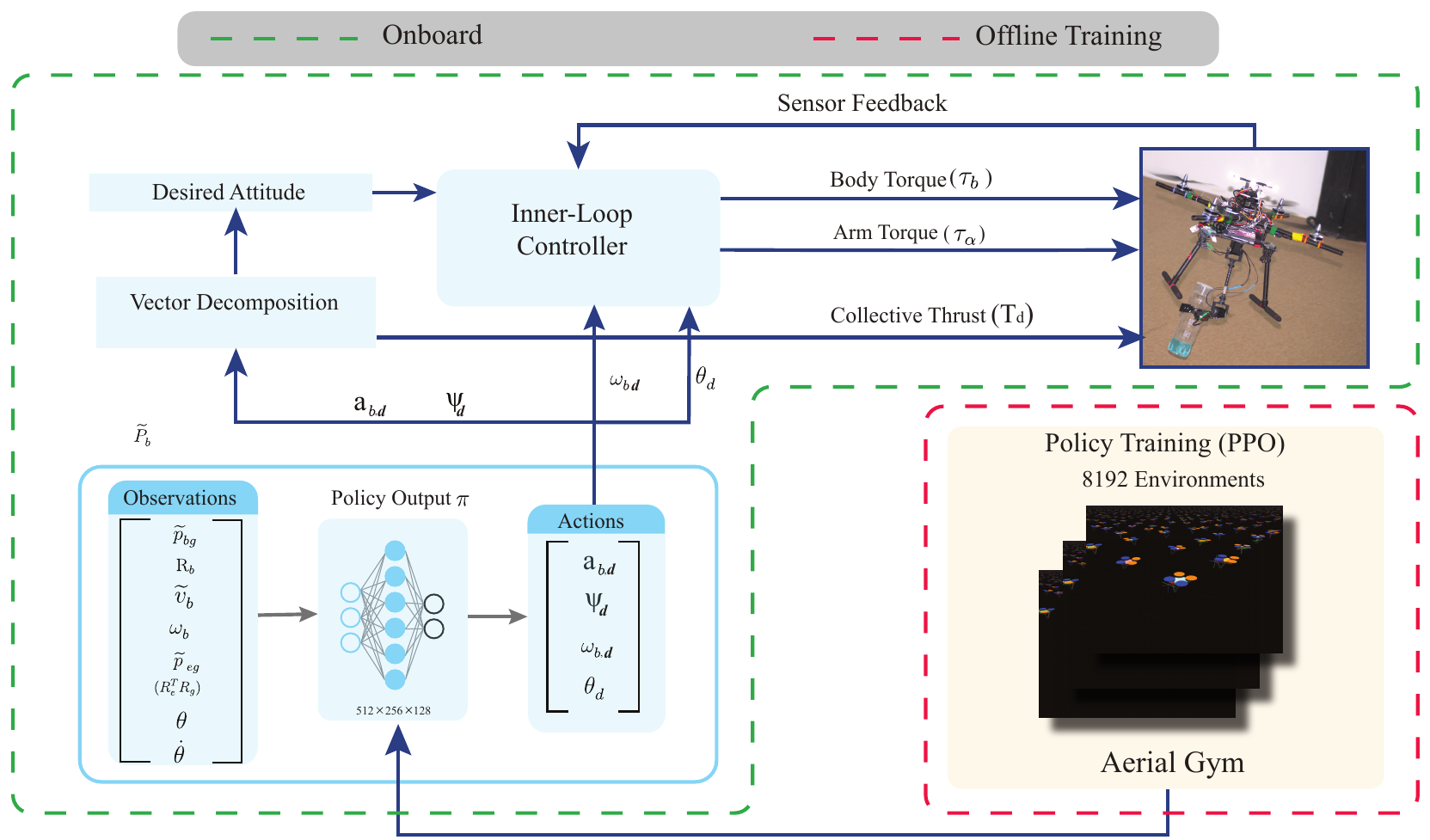}
    \caption{Overview of the proposed hierarchical control framework.}
    \vspace{-2mm}
    \label{fig:Pipeline}
\end{figure*}

% \subsection{Contributions}
Despite substantial progress, direct whole-body aerial manipulation under abrupt payload changes, arm-induced motion, and dynamic uncertainty remains an open challenge. We address this with a hierarchical RL framework with inner-loop uncertainty estimation for aerial manipulators (Fig.~\ref{fig:Pipeline}). The RL policy maps a desired 6-DOF end-effector target to coordinated whole-body commands, while the inner-loop controller tracks these commands and compensates for transient unknown dynamics during execution without requiring an a priori system model.
The main contributions are:
\begin{itemize}
\item A hierarchical architecture combining RL-based whole-body command generation with estimator-based inner-loop tracking and uncertainty compensation.
\item A dynamics-estimation execution layer for transient UAV--arm coupling changes, including grasp/release-induced inertial variation and rapid manipulator motion.
\item Evaluation under off-nominal aerial-manipulation conditions, including payload variation and dynamic coupling during arm motion.
\item Demonstration that learned coordination with estimator-based feedback improves end-effector regulation accuracy and smoothness under dynamic uncertainty.
\end{itemize}

% While the state-of-the-art have advanced the field substantially, the combined problem of direct whole-body aerial manipulation under abrupt payload variation, arm motion, and related dynamic uncertainties remain a significant open challenge. In this direction, we propose a hierarchical RL-based framework with inner-loop uncertainty estimation for aerial manipulators (cf. Fig.~\ref{fig:Pipeline}): the RL generates coordinated whole-body commands from a desired 6-DOF end-effector target, while an inner loop control tracks these commands and compensates for transient unknown dynamic uncertainty during execution without a priori knowledge of system model.
% The main contributions of this paper are as follows:
% \begin{itemize}
%   \item We develop a hierarchical control architecture that combines RL-based whole-body command generation with an estimator-based inner-loop controller for tracking and uncertainty compensation in aerial manipulators.
%   \item We introduce a dynamics estimator execution layer to compensate for transient changes in the coupled UAV--arm dynamics, including grasp/release-induced inertial variation and rapid manipulator motion.
%   \item We evaluate the framework under off-nominal aerial-manipulation conditions, including payload variation and off-nominal dynamic coupling during manipulator motion.
%   \item We show that combining learned coordination with estimator-based feedback improves end-effector regulation accuracy and smoothness under dynamic uncertainty.
% \end{itemize}

\section{Proposed Methodology}

\subsection{Notation and Preliminaries}
\label{sec:notation}

We define four primary coordinate frames: the inertial World frame $\{W\}$, the quadrotor Body frame $\{B\}$ attached to its center of mass, the End-effector (gripper) frame $\{E\}$ attached to the geometric center of the gripper claws and the goal frame $\{G\}$. The system motion variables are denoted as follows: linear (position, velocity, acceleration) as $(\mathbf{p},\mathbf{v},\mathbf{a})$, quaternion as $\mathbf{q}$, Euler angles (roll, pitch, yaw) as $(\phi,\vartheta,\psi)$, angular velocity as $\boldsymbol{\omega}$; Rotation matrix as $\mathbf{R}$ and arm joint angular position and velocity as $(\boldsymbol{\theta},\dot{\boldsymbol{\theta}})$. The symbol $t$ denotes time, while $T$ denotes collective thrust. The inner loop control output (torque) is denoted by $\tau$. A leading superscript denotes the reference frame in which a physical quantity is expressed. The following subscript denotes the entity for which the physical quantity is expressed ($b$ for aerial base, $e$ for end effector, $\alpha$ for manipulator arm and $g$ for target). The subscript $d$ or $\{(\cdot),d\}$ represents a desired quantity either outputted by RL agent or the controller. For example, ${}^W\mathbf{p}_e \in \mathbb{R}^3$ represents the position of the end effector expressed in the world frame and ${}^B\boldsymbol{\omega}_{b,d} \in \mathbb{R}^3$ represents the desired angular velocity of the aerial base in the body frame. The rotation matrix ${}^W\mathbf{R}_{B} \in SO(3)$ and the quaternion ${}^W\mathbf{q}_B$ maps vectors from $\{B\}$ to $\{W\}$.

\subsection{System Overview}

We consider an aerial manipulator composed of a quadrotor and a $3$-DoF arm. The objective is goal-conditioned end-effector pose regulation: given a desired target pose $({}^W\mathbf{p}_g,{}^W\mathbf{R}_G)\in SE(3)$, the controller must generate coordinated base and manipulator motion that drives the end effector to the goal while preserving stable flight. The overall architecture is hierarchical. A forward-kinematics block reconstructs the end-effector pose from the measured base state and joint configuration, an RL policy generates high-level motion commands, and an inner-loop controller that converts these commands into actuator inputs.

At each control step, the policy receives the current base pose, base velocity, body angular velocity, end-effector pose, and joint states, together with the target pose. The policy then outputs desired base linear acceleration, desired yaw, desired body rates, and desired joint-position commands. The desired linear acceleration and yaw are converted into a desired base attitude and the required collective thrust via thrust vector decomposition, while the desired body rates serve as rotational feedforward commands. Finally, these reference commands, together with real-time attitude and body-rate measurements from the flight controller, are fed into the estimator-based inner-loop controller. This controller computes the body torques required for the aerial base and the joint torques required for the manipulator arm.

{ \small
\begin{equation}
\label{eq:closed_loop_map}
\mathbf{o}(t)=\mathcal{O}(\mathbf{x}(t),\mathcal{G}),\quad
\mathcal{A}(t)=\pi(\mathbf{o}(t)),\quad
\mathbf{u}(t)=
\big[
\boldsymbol{\tau}_{b}^{\top}(t)\ 
\boldsymbol{\tau}_{\alpha}^{\top}(t)\ 
T_d(t)
\big]^{\top}
\end{equation}
}

where $\mathbf{x}(t)$ denotes the full system state, $\mathcal{G} = [{}^W\mathbf{p}_g^\top,\, ({}^W\phi_{g},{}^W\vartheta_{g},{}^W\psi_{g})^\top]^\top$ represents the 6 DOF goal pose, $\mathbf{o}(t)$ is the policy observation, $\mathcal{A}(t)$ is the high-level actions, and $\mathbf{u}(t)$ collects the low-level control inputs applied to the system. The system evolves as $\mathbf{x}(t+1)=f(\mathbf{x}(t),\mathbf{u}(t))$.

\subsection{Reinforcement Learning Formulation}

We model the problem as a continuous-state, continuous-action Markov decision process. The observation vector has dimension $33$ and is constructed as
\begin{equation}
\label{eq:obs_vec}
\begin{aligned}
\mathbf{o}(t)=
\big[
&{}^W\tilde{\mathbf{p}}_{gb}^{\top},
\mathrm{vec}\!\left({}^W\mathbf{R}_B(:,0\!:\!2)\right)^{\top},
{}^W\tilde{\mathbf{v}}_{b}^{\top},
{}^B\boldsymbol{\omega}_b^{\top},\\
&{}^W\tilde{\mathbf{p}}_{ge}^{\top},
\mathrm{vec}\!\left({}^W\mathbf{R}_E^{\top}{}^W\mathbf{R}_G\right)^{\top},
\boldsymbol{\theta}^{\top},
\dot{\boldsymbol{\theta}}^{\top}
\big]^{\top}\in\mathbb{R}^{33},
\end{aligned}
\end{equation}
where
\begin{equation}
\label{eq:obs_norm}
{}^W\tilde{\mathbf{p}}_{gb}=
\frac{{}^W\mathbf{p}_g-{}^W\mathbf{p}_b}{10},\quad
{}^W\tilde{\mathbf{v}}_{b}=
\frac{{}^W\mathbf{v}_b}{10},\quad
{}^W\tilde{\mathbf{p}}_{ge}=
\frac{{}^W\mathbf{p}_g-{}^W\mathbf{p}_e}{10}. \nonumber
\end{equation}

The observation design exposes both task-space geometric errors and motion-state variables required for stable whole-body coordination. The action vector is given by
\begin{equation}
\label{eq:actions}
\mathcal{A}(t)=
\big[
({}^W\mathbf{a}_{b,d})^\top,
({}^B\boldsymbol{\omega}_{b,d})^\top,
\psi_d,
\boldsymbol{\theta}_d^\top
\big]^\top\in \mathbb{R}^{10},
\end{equation}
with $3$ translational components, $3$ body-rate components, $1$ yaw command, and $3$ desired joint positions. This action parameterization is deliberate: the policy learns task-level coordinated motion commands, while the inner loops handle UAV attitude tracking and manipulator joint actuation.

\subsection{Reward Function Design}

We use a two-scale exponential shaping for rewards with a general structure as in \eqref{eq:rew_gen}. We superimpose two exponential functions, one of wider variance and another of lower variance. The higher variance exponential prevents gradients from vanishing when the system is away from the target, whereas the lower variance exponential maintains high reward gradients near the target enforcing precision. The penalties are defined by \eqref{eq:penalty_gen}, which is always negative.

\begin{align}
&r = r_1\,e^{-r_2\,x^2}
+
r_3\,e^{-r_4\,x^2} \label{eq:rew_gen}\\
& p = p_1(e^{-p_2\,x^2}-1) \label{eq:penalty_gen}
\end{align}

\begin{table}[t]
\caption{Reward and Penalty Configurations}
\label{tab:rewards_penalties}
\centering
\scriptsize
\renewcommand{\arraystretch}{0.90}
\setlength{\tabcolsep}{1.8pt}
\makebox[\columnwidth][c]{%
\scalebox{0.90}{%
\begin{tabular}{@{}lccccp{0.30\columnwidth}@{\hspace{3mm}}lccp{0.36\columnwidth}@{}}
\toprule
\multicolumn{6}{c}{\textbf{Rewards}} &
\multicolumn{4}{c}{\textbf{Penalties}} \\
\cmidrule(lr){1-6} \cmidrule(lr){7-10}
Term & $r_1$ & $r_2$ & $r_3$ & $r_4$ & $x$ &
Term & $p_1$ & $p_2$ & $x$ \\
\midrule
$r_{\mathrm{pos}}$ & 10 & 10 & 2 & 2 &
${}^W\mathbf{p}_g-{}^W\mathbf{p}_e$
&
$r_{\mathrm{\|a\|}}$ & 4 & 1 &
${}^W\mathbf{a}_{b,d}$ \\

$r_{\mathrm{ori}}$ & 20 & 5 & 4 & 1 &
$2\arccos\!\bigl(\lvert{}^W\mathbf{q}_E^\top{}^W\mathbf{q}_G\rvert\bigr)$
&
$r_{\mathrm{\|\omega\|}}$ & 4 & 5 &
${}^B\boldsymbol{\omega}_{b,d}$ \\

$r_{\mathrm{linvel}}$ & 1 & 5 & 0 & 0 &
${}^W\mathbf{v}_b$
&
$r_{\mathrm{\|\psi\|}}$ & 2 & 5 &
$\psi_d$ \\

$r_{\mathrm{angvel}}$ & 1.5 & 2 & 0 & 0 &
${}^B\boldsymbol{\omega}_b$
&
$r_{\mathrm{\Delta a}}$ & 4 & 5 &
${}^W\mathbf{a}_{b,d}(t)-{}^W\mathbf{a}_{b,d}(t-1)$ \\

$r_{\mathrm{jvel}}$ & 2 & 5 & 0 & 0 &
$\dot{\boldsymbol{\theta}}$
&
$r_{\mathrm{\Delta \omega}}$ & 4 & 12 &
${}^B\boldsymbol{\omega}_{b,d}(t)-{}^B\boldsymbol{\omega}_{b,d}(t-1)$ \\

$r_{\mathrm{eevel}}$ & 0.5 & 5 & 0.5 & 2 &
${}^W\mathbf{v}_e$
&
$r_{\mathrm{\Delta \psi}}$ & 4 & 12 &
$\psi_d(t)-\psi_d(t-1)$ \\

$r_{\mathrm{up}}$ & 2.5 & 5 & 2.5 & 2 &
$1-e_3^\top{}^W\mathbf{R}_B e_3$
&
$r_{\mathrm{\Delta \theta}}$ & 4 & 12 &
$\boldsymbol{\theta}_d(t)-\boldsymbol{\theta}_d(t-1)$ \\
\bottomrule
\end{tabular}%
}}
\vspace{-1mm}

% {\tiny Reward terms use \eqref{eq:rew_gen}, and penalty terms use \eqref{eq:penalty_gen}.}
\vspace{-2mm}
\end{table}

% \begin{table}[t]
% \footnotesize
% \renewcommand{\arraystretch}{1.2}
% \caption{\small Tracking performance comparison (RMS)}
% % \vspace{-1.5mm}
% 		\centering
% {
% \scalebox{0.8}{	
% \begin{tabular}{c|ccc|ccc|cc}
% \hline
% \multirow{2}{*}{} & \multicolumn{3}{c|}{Position Error (m)}                                   & \multicolumn{3}{c|}{Attitude Error (deg)}     
% & \multicolumn{2}{c}{Arm Error (deg)}
%       \\ \cline{2-9} 
%                   & $x$  & $y$  & $z$        & $\phi$ & $\theta$  & $\psi$   & $\alpha_1$         & $\alpha_2$         \\ \hline
% DOC   & $0.093 $& $0.041 $& $0.153$& $1.603$& $5.142$& $1.987$& $3.881$& $3.091$\\ \hline
% NS-AC   & $0.079 $& $0.038$& $0.105 $& $1.477$& $4.097$& $1.128$& $2.225 $& $2.016$\\ \hline
% Proposed   & $\mathbf{0.031}$& $\mathbf{0.019}$& $\mathbf{0.052}$& $\mathbf{1.128 }$& $\mathbf{2.018}$& $\mathbf{1.033 }$& $\mathbf{1.107}$& $\mathbf{1.015}$\\ \hline
% \end{tabular}}}
% \label{table_performance}

% \end{table}

We reward the agent for reaching the goal pose (position reward $r_{\mathrm{pos}}$, orientation reward $r_{\mathrm{ori}}$), for maintaining low velocities (linear velocity reward $r_{\mathrm{linvel}}$, angular velocity reward $r_{\mathrm{angvel}}$, joint angular velocity reward $r_{\mathrm{jvel}}$, end-effector velocity reward $r_{\mathrm{eevel}}$) and for maintaining upright position of aerial base (upright reward $r_{\mathrm{up}}$). We penalize it for high action magnitudes (linear acceleration penalty $r_{\mathrm{\|a\|}}$, body rate penalty $r_{\mathrm{\|\omega\|}}$, yaw penalty $r_{\mathrm{\|\psi\|}}$) and for high difference between current and previous actions (linear acceleration difference penalty $r_{\mathrm{\Delta a}}$, body rate difference penalty $r_{\mathrm{\Delta \omega}}$, yaw difference penalty $r_{\mathrm{\Delta \psi}}$, joint action difference penalty $r_{\mathrm{\Delta \theta}}$), to enforce smoothness. The coefficients for the rewards and penalties are given in Table~\ref{tab:rewards_penalties}.

We also introduce a temporal shaping reward (progress reward) shown in equation \eqref{eq:eqr_prog}. It rewards progress of the end effector towards the goal position. The asymmetry of this reward, i.e. a higher penalty for moving away strongly discourages limit cycles and oscillatory behavior.
\begin{equation}
\label{eq:eqr_prog}
\begin{gathered}
r_{\mathrm{prog}}=
\begin{cases}
50\,\Delta e_p, & \Delta e_p>0,\\
100\,\Delta e_p, & \Delta e_p\le 0,
\end{cases}\\[-1mm]
\Delta e_p =
\|{}^W\mathbf{e}_{ge}(t-1)\|_2-\|{}^W\mathbf{e}_{ge}(t)\|_2,\quad
{}^W\mathbf{e}_{ge} = {}^W\mathbf{p}_g-{}^W\mathbf{p}_e .
\end{gathered}
\end{equation}

The reward components are aggregated as
\begin{equation}
\label{eq:rgeneral}
\begin{aligned}
r_{\mathrm{general}}
&= k_{pos}\,r_{\mathrm{pos}} + k_{ori}\,r_{\mathrm{ori}}\\
&\quad + k_{pen}(\,r_{\Delta a} + \,r_{\Delta \omega} + \,r_{\Delta \psi} + \,r_{\Delta \theta} \\
&\quad + \,r_{\|a\|} + \,r_{\|\omega\|} + \,r_{\|\psi\|}) + r_{\mathrm{up}}\\
&\quad + r_{\mathrm{linvel}} + r_{\mathrm{angvel}}  + r_{\mathrm{eevel}} + r_{\mathrm{jvel}},
\end{aligned}
\end{equation}
\begin{equation}
\label{eq:rnear}
\begin{aligned}
r_{\mathrm{near}}
&= \,r_{\mathrm{ori}} + r_{\mathrm{linvel}} + r_{\mathrm{angvel}}  + r_{\mathrm{eevel}}\\
&\quad + r_{\mathrm{jvel}} + \,r_{\|a\|} + \,r_{\|\omega\|} + \,r_{\|\psi\|}\\
&\quad + \,r_{\Delta a} + \,r_{\Delta\omega} + \,r_{\Delta \psi} + \,r_{\Delta \theta},
\end{aligned}
\end{equation}
and the total reward is defined as
\begin{equation}
r_{total}
=
r_{\mathrm{prog}}
+
\frac{r_{\mathrm{pos}}\,r_{\mathrm{near}} + r_{\mathrm{general}}}{100}.
\end{equation}

The weighting coefficients $k_{pos}$, $k_{ori}$, and $k_{pen}$ balance the positional, orientation, and penalty terms. We set them to $k_{pos}=50$, $k_{ori}=3$, and $k_{pen}=2.5$.

To ensure a robust learning process and prevent reward hacking, we employ a \textit{multiplicative coupling} between the primary task objective and secondary performance metrics. Specifically, the reward structure integrates the positional accuracy ($r_{\text{pos}}$) with stability and smoothness objectives ($r_{\text{near}}$) such that the agent can only ``unlock'' significant secondary rewards upon satisfying the primary goal of reaching the target. This hierarchical reward shaping ensures that the policy prioritizes global convergence before optimizing for fine-grained control effort and stability.

\subsection{Policy Training Procedure and Setup}

The policy is trained in a GPU-accelerated vectorized simulation on an NVIDIA GeForce RTX~4090, using the same structured closed loop as the real system in Fig.~\ref{fig:Pipeline}. At each step, the simulator provides $\mathbf{x}(t)$, the observation $\mathbf{o}(t)$ is constructed, and the policy outputs $\mathcal{A}(t)$. This action is converted into desired attitude, thrust, and joint references, which are tracked by the estimator-based inner loop and joint controller. Thus, training occurs with the same control structure used during deployment rather than a simplified kinematic abstraction.

We use PPO~\cite{schulman2017proximal} through RL Games in Aerial Gym~\cite{kulkarni2025aerial}. The actor and critic are MLPs with hidden layers $[512,256,128]$ and ELU activations. Training uses $8192$ parallel environments, horizon length $32$, minibatch size $8192$, $4$ epochs per update, discount factor $0.99$, learning rate $10^{-4}$ with adaptive scheduling, KL threshold $0.008$, and clipping parameter $\epsilon=0.1$. The RL loop runs at $100$~Hz with ONNX Runtime inference, while the inner-loop controller runs at $300$~Hz. Episodes terminate after 500 steps or when the end-effector--goal distance exceeds $5.5$\,m.

For robustness, domain randomization is applied at every reset: payload mass is sampled from $[0.1,0.5]$~kg, manipulator link lengths from $[0.8,1.2]$ with consistent inertial scaling, motor time constants from $[0.95,1.05]$, joint stiffness from $[0.5,1.5]$, and joint damping from $[0.9,1.1]$. Initial poses and velocities are randomized, and target orientation is sampled with roll/pitch in $\left[-\frac{\pi}{6},\frac{\pi}{6}\right]$ and yaw in $\left[-\pi,\pi\right]$.

\subsection{Control Architecture}

The inner-loop controller receives the policy output $\mathcal{A}(t)$ from \eqref{eq:actions} and generates the corresponding low-level control input $\tau$. The aerial-base branch converts the commanded acceleration and yaw $({}^W\mathbf{a}_{b,d},\psi_d)$ into the desired attitude ${}^W\mathbf{R}_{B,d}$ and collective thrust $T_d$ using the standard geometric quadrotor construction of Mellinger and Kumar~\cite{mellinger2011minimum}. The corresponding attitude and angular-rate tracking errors are also computed following~\cite{mellinger2011minimum}, and are passed to the estimator-based inner-loop controller together with the desired joint configuration $\boldsymbol{\theta}_d$.

% Therefore, the RL policy operates at the motion-reference level rather than directly issuing actuator commands. Specifically, it outputs a desired translational acceleration, a desired yaw angle, and a desired body-rate reference. Through vector decomposition, the desired translational acceleration and yaw are converted into the desired vehicle attitude ${}^W\mathbf{R}_{B,d}$ and the collective thrust command. The estimator-based inner-loop controller then uses $(\phi_{b,d},\vartheta_{b,d},\psi_{b,d}) = \mathrm{Euler}({}^W\mathbf{R}_{B,d})$, the desired body-rate reference ${}^B\boldsymbol{\omega}_{b,d}$ as rotational feedforward information, and the desired joint configuration $\boldsymbol{\theta}_d$, together with the measured vehicle attitude $(\phi_{b},\vartheta_{b},\psi_{b}) = \mathrm{Euler}({}^W\mathbf{R}_{B})$, measured body rates ${}^B\boldsymbol{\omega}_b$, and manipulator states, to compute the low-level control inputs, namely the quadrotor body torques and manipulator joint torques. In this way, the proposed architecture maintains a clear and physically grounded separation between high-level motion generation and low-level coupled stabilization and tracking.

Therefore, the RL policy operates at the motion-reference level rather than directly issuing actuator commands. Specifically, it outputs a desired translational acceleration, a desired yaw angle, and a desired body-rate reference. Through vector decomposition, the desired translational acceleration and yaw are converted into the desired vehicle attitude ${}^W\mathbf{R}_{B,d}$ and the collective thrust command. The estimator-based inner-loop controller then uses $(\phi_{b,d},\vartheta_{b,d},\psi_{b,d}) = \mathrm{Euler}({}^W\mathbf{R}_{B,d})$, the desired body-rate reference ${}^B\boldsymbol{\omega}_{b,d}$ as rotational feedforward information, and the desired joint configuration $\boldsymbol{\theta}_d$, together with the measured vehicle attitude $(\phi_{b},\vartheta_{b},\psi_{b}) = \mathrm{Euler}({}^W\mathbf{R}_{B})$, measured body rates ${}^B\boldsymbol{\omega}_b$, and manipulator states, to compute the low-level control inputs, namely the quadrotor body torques and manipulator joint torques. In this way, the proposed architecture maintains a clear and physically grounded separation between high-level motion generation and low-level coupled stabilization and tracking.

\subsection{Inner-Loop Controller}

To describe the controller, we define a generic configuration vector $\eta=[(\phi_{b},\vartheta_{b},\psi_{b})^\top,\boldsymbol{\theta}^\top]^\top$. The coupled inner-loop dynamics in the standard Euler-Lagrange form can then be written as~\cite{ollero2022past,orsag2017dexterous}
\begin{equation}
M_\eta (\eta) \ddot{\eta} + H_\eta (\eta, \dot{\eta}, t) = \tau_\eta,
\label{eq:dyn_true}
\end{equation}
where $M_\eta$ is the effective inertia matrix, $H_\eta$ collects the Coriolis, dissipative, coupling, and disturbance terms, and $\tau_\eta$ is the torque input. Introducing a user-defined constant diagonal matrix $\bar M_\eta$, the same dynamics can be rewritten as
\begin{align}
& \bar M_\eta \ddot{\eta} + \bar H_\eta(\eta,\dot{\eta},t) = \tau_\eta,
\label{eq:dyn_nom} \\
\text{with}~~ & \bar H_\eta = (M_\eta-\bar M_\eta)\ddot{\eta}+H_\eta
\end{align}
being defined as the lumped unknown dynamics and uncertainty. Note that the coupled inertial terms, represented by the off-diagonal terms in $M_\eta$ and other terms $H_\eta$, are the ones which are the most difficult to model. Hence, $\bar M_\eta$ is selected as a diagonal matrix for control design, whereas the off-diagonal terms are grouped under unknown lumped uncertainty function $\bar H_\eta$.

Let $\eta_d$ denote the desired generalized-coordinate trajectory induced by $((\phi_{b,d},\vartheta_{b,d},\psi_{b,d})^\top, \boldsymbol{\theta}_d^\top)$, and define the tracking error as $e_\eta=\eta-\eta_d$ and $\dot e_\eta=\dot\eta-\dot\eta_d$. The estimator-based inner-loop control law is defined as
\begin{equation}
\label{eq:ctrl_law}
\tau_\eta =
\bar M_\eta
\left(
\ddot{\eta}_d
-K_{p\eta}e_\eta
-K_{d\eta}\dot e_\eta
\right)
+\hat{\bar H}_\eta .
\end{equation}
where $K_{p\eta}$ and $K_{d\eta}$ are positive-definite gain matrices and $\hat{\bar H}_\eta$ is a delay-based estimate of the lumped uncertainty computed from past state and control input as
\begin{equation}
\hat{\bar H}_\eta(t)
=
\tau_\eta(t-L)-\bar M_\eta \ddot{\eta}(t-L),
\label{eq:hhat}
\end{equation}
with $L>0$ is the delay induced owing to usage of past data.
Substituting \eqref{eq:ctrl_law} into \eqref{eq:dyn_nom} yields the closed-loop error dynamics
\begin{equation}
\ddot e_\eta + K_{d\eta}\dot e_\eta + K_{p\eta} e_\eta
=
\bar M_\eta^{-1}\!\left(\hat{\bar H}_\eta-\bar H_\eta\right).
\label{eq:error_dyn}
\end{equation}
From the properties of Euler--Lagrange robotic system dynamics it is well known that they do not exhibit finite-time escape (cf. \cite[Ch. 2]{ortega1998euler}). Therefore, the mismatch $(\hat{\bar H}_\eta-\bar H_\eta)$, i.e., the discrepancy in state evaluation between the delay interval $L$ always remains bounded (cf. \cite[Ch. 3]{khalil2002nonlinear}). Hence, there exists a positive scalar $\epsilon<\infty$ such that
\begin{equation}
\left\|
\bar M_\eta^{-1}\!\left(
\hat{\bar H}_\eta-\bar H_\eta
\right)
\right\|
\le \epsilon .
\label{eq:bounded_rhs}
\end{equation}
Therefore, the right-hand side of \eqref{eq:error_dyn} is bounded, and the tracking error remains bounded by the standard results for linear systems with bounded input. This boundedness argument justifies the use of the estimator-based inner loop for compensating modeling mismatch, unmodeled coupling, and fast transient variations during flight.

We implement the inner-loop controller as the torque-level tracking layer for both the aerial base and the manipulator. Following the formulation in this subsection, the controller takes the desired attitude $(\phi_{b,d},\vartheta_{b,d},\psi_{b,d})$, and desired joint configuration $\boldsymbol{\theta}_d$ as reference inputs, while the RL-generated body-rate commands are incorporated as feedforward rotational terms. The loop is closed using real-time measurements of the vehicle attitude, body rates, and manipulator joint positions from the onboard flight controller sensors and the Dynamixel encoders, respectively. Based on these signals, the controller computes the body-torque command $\tau_b$ for the quadrotor and the joint-torque command $\tau_\alpha$ for the manipulator. In practice, direct low-level torque interfacing with the Dynamixel actuators is nontrivial; therefore, we transmit $\tau_\alpha$ through the ROS~2 control stack, which provides reliable actuator communication and executes the command in current-based torque mode. Importantly, the proposed control law does not require a fully accurate coupled system model. The delay parameter $L$ is selected as the smallest interval over which past measurements are available, which is naturally determined by the control update rate. Likewise, $\bar{M}_{\eta}$ is chosen using nominal diagonal mass/inertia terms, while the remaining coupling effects, unmodeled dynamics, and disturbances are compensated through the delay-based estimate $\hat{\bar H}_\eta(t)$.

\begin{figure*}[h]
  \centering
\includegraphics[width=0.66\textwidth]{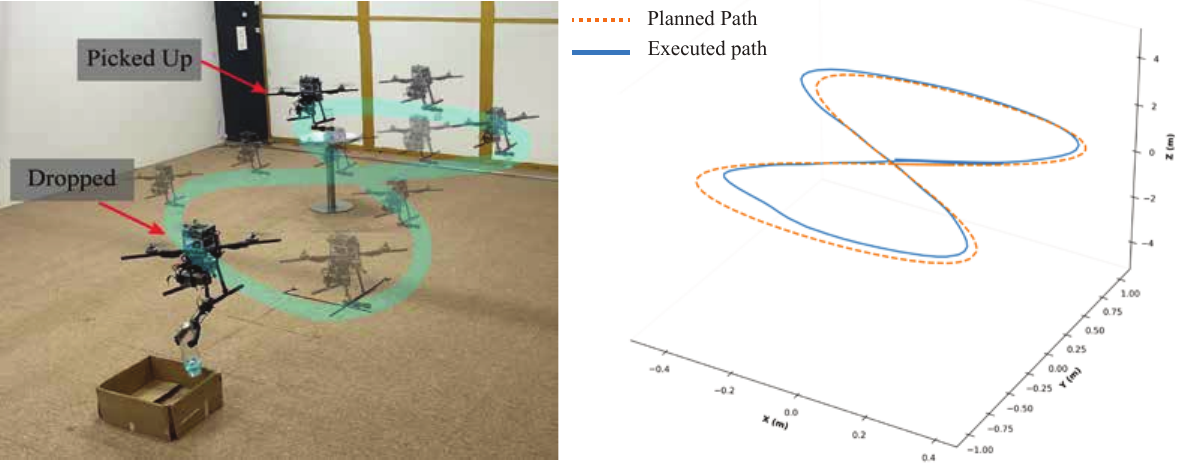}
  \caption{Hardware trajectory-tracking result for the figure-eight experiment, with annotated pick-up and drop-off locations indicating the grasp and release events that induce in-flight payload variation.}
  \vspace{-2mm}
  \label{fig:trajectory}
\end{figure*}

\section{Experiments}

We validate the proposed framework on an aerial-manipulation platform and evaluate three questions: 1) whether the method can accurately track planned path in hardware, 2) how performance changes under increased payload and speed, and 3) how it compares with other baselines.

\begin{figure}[t]
    \centering
    \includegraphics[width=0.34\textwidth]{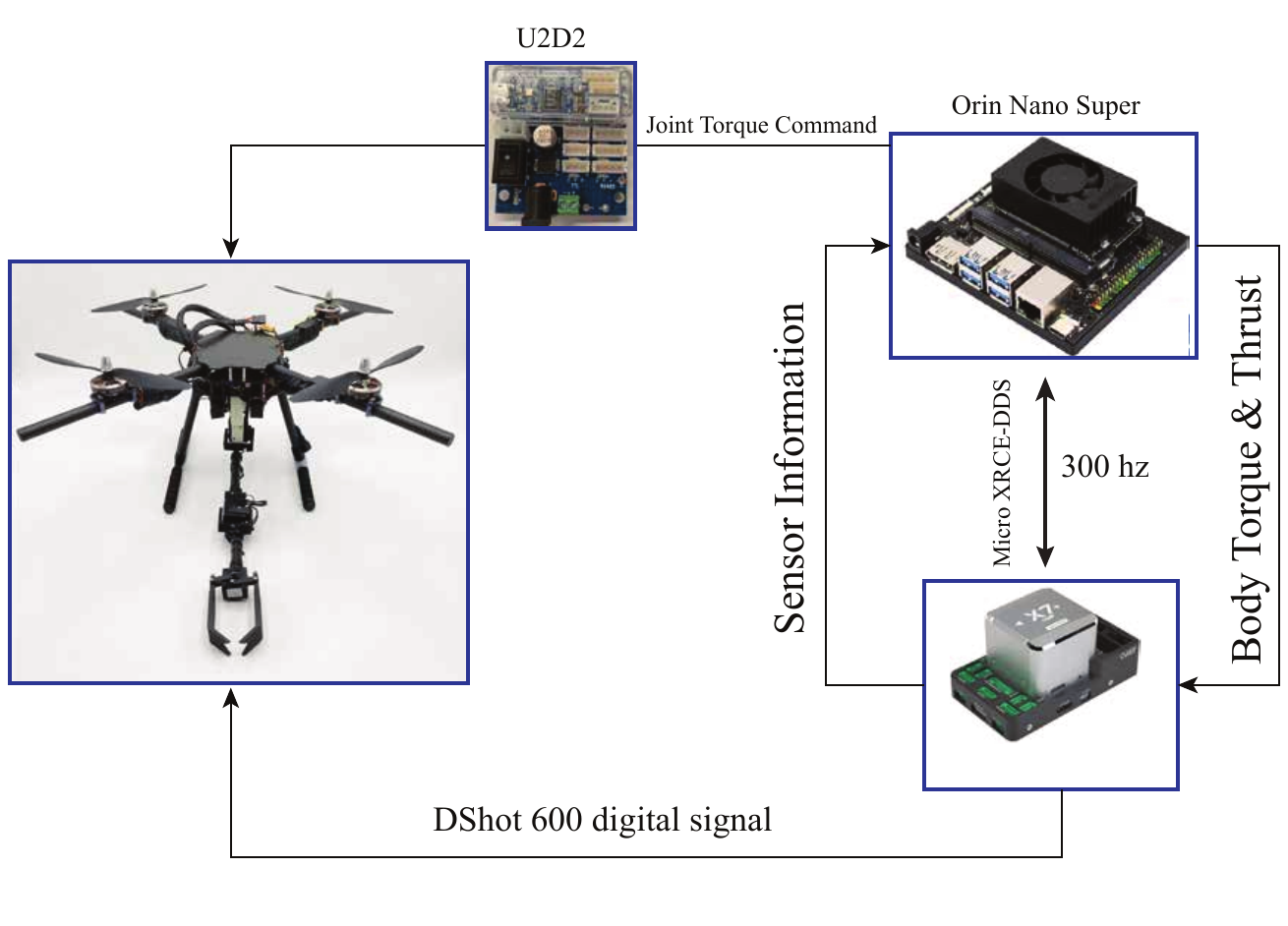}
    \caption{Experimental platform and onboard control architecture.}
    \vspace{-2mm}
    \label{fig:hardware}
\end{figure}

% \subsection{Experimental Setup}

% \paragraph{Hardware Setup}
% For hardware validation, we use a Tarot-450 quadrotor equipped with SunnySky V4006 motors, 13-inch propellers, a 6S LiPo battery, and a 3-DoF manipulator actuated by Dynamixel XM430-W210-T motors. A custom 3D-printed gripper, driven by an additional Dynamixel actuator, is used for pick-and-place experiments. The complete platform weighs approximately $3.0$~kg, and the manipulator and gripper are powered through a U2D2 Power Hub Board, as shown in Fig.~\ref{fig:hardware}.

% The aerial base runs customized PX4 firmware on a CUAV X7+ flight controller, while high-level computation, policy inference, and the estimator-based inner loop run onboard a Jetson Orin Nano Super. The Jetson communicates with the flight controller through \textit{micro-ROS}/\textit{DDS} in \textit{ROS~2}. Manipulator commands are sent through the \textit{ROS2\_control} Joint Trajectory Controller in current-based torque mode; since the Dynamixel actuators do not provide direct torque sensing, commanded torques are approximated using internal effort estimation and tracked at 100~Hz.

% Quadrotor state feedback is provided by an OptiTrack system at 120~Hz fused with onboard IMU measurements, while manipulator joint positions and velocities are obtained from Dynamixel encoders, with joint accelerations computed numerically. The learned policy and estimator-based controller generate collective thrust and body/joint torque commands at 100~Hz, and the customized PX4 stack converts these commands into rotor-speed commands at 300~Hz.

\subsection{Experimental Setup}

\paragraph{Hardware Setup}
Hardware experiments are conducted on a Tarot-450 quadrotor with SunnySky V4006 motors, 13-inch propellers, a 6S LiPo battery, and a 3-DoF Dynamixel XM430-W210-T manipulator. A custom 3D-printed gripper driven by an additional Dynamixel actuator is used for pick-and-place tasks. The complete platform weighs approximately $3.0$~kg, with the manipulator and gripper powered through a U2D2 Power Hub Board (Fig.~\ref{fig:hardware}).

A CUAV X7+ flight controller runs customized PX4 firmware for the aerial base, while a Jetson Orin Nano Super handles high-level computation, policy inference, and the estimator-based inner loop. Communication between the Jetson and flight controller is handled through \textit{micro-ROS}/\textit{DDS} in \textit{ROS~2}. Manipulator commands are sent through \textit{ROS2\_control} in current-based torque mode, with commanded torques approximated from Dynamixel internal effort estimates and tracked at 100~Hz.

Base state feedback comes from OptiTrack at 120~Hz fused with onboard IMU data, while manipulator joint states are obtained from Dynamixel encoders. The learned policy and estimator-based controller run at 100~Hz and generate collective thrust and body/joint torque commands, which are converted by the customized PX4 stack into rotor-speed commands at 300~Hz.

\subsection{Experimental Protocol}

We evaluate the platform on a planar figure-eight trajectory, which induces repeated curvature reversals and strong translational--rotational coupling. The trajectory is executed with payloads of 200~g and 400~g at nominal speeds of 0.5~m/s and 1~m/s. During execution, the platform also performs pick-up and drop-off actions at the locations marked in Fig.~\ref{fig:trajectory}, creating in-flight payload changes and testing grasp/release-induced transient mismatch. All controllers are evaluated under the same trajectory, payload, and speed settings, and the resulting end-effector tracking performance is summarized in Table~\ref{tab:baseline_comparison}.

\subsection{Baselines and Evaluation Metrics}

We compare against two baselines with the same RL outer loop but different inner-loop controllers: RL+PID, using cascaded PID for the aerial base and joint-space PID for the arm, and RL+INDI+PID, using INDI for the base while retaining PID arm tracking. Performance is measured using end-effector tracking RMSE and task success rate under the tested high-speed and high-payload conditions.

\section{Results and discussion}

\begin{table}[t]
\centering
\caption{Baseline comparison on the figure-eight path. Each entry reports RMSE [m] / success rate [\%].}
\label{tab:baseline_comparison}
\vspace{-1mm}
\scriptsize
\renewcommand{\arraystretch}{0.88}
\setlength{\tabcolsep}{2.0pt}
\makebox[\columnwidth][c]{%
\scalebox{0.92}{%
\begin{tabular}{@{}ccccc@{}}
\toprule
Payload & Speed & RL+PID & INDI & Ours \\
\midrule
200 g & 0.5 m/s & 0.242 / 85 & 0.195 / 95 & 0.180 / 95 \\
200 g & 1.0 m/s & 0.315 / 70 & 0.220 / 90 & 0.151 / 95 \\
400 g & 0.5 m/s & 0.288 / 75 & 0.210 / 95 & 0.195 / 95 \\
400 g & 1.0 m/s & 0.392 / 60 & 0.285 / 85 & 0.205 / 90 \\
\midrule
RMSE std. [m] & -- & 0.062 & 0.038 & 0.024 \\
RMSE red. [\%] & -- & 0.0 & 26.4 & 40.9 \\
\bottomrule
\multicolumn{5}{@{}l}{\scriptsize INDI denotes the RL + INDI + PID arm baseline.}
\end{tabular}%
}}
\vspace{-2mm}
\end{table}

As shown in Fig.~\ref{fig:hardware} and Fig.~\ref{fig:trajectory}, the proposed controller is deployable on hardware and achieves stable coupled flight--manipulation. The executed figure-eight trajectory closely tracks the planned path, with bounded errors during simultaneous base motion and arm actuation. Table~\ref{tab:baseline_comparison} shows that the proposed method attains the lowest end-effector RMSE across all payload--speed conditions, with the largest gains in aggressive cases such as \(200\,\mathrm{g}\), \(1\,\mathrm{m/s}\) and \(400\,\mathrm{g}\), \(1\,\mathrm{m/s}\). The success rates show the same trend, indicating improved reliability as payload and speed increase.
Aggregated over the hardware trials, the proposed method reduces mean RMSE by \(40.9\%\) relative to RL+PID and \(26.4\%\) relative to RL+INDI+PID. It also lowers the RMSE standard deviation to \(0.024\,\mathrm{m}\), compared with \(0.038\,\mathrm{m}\) for RL+INDI+PID and \(0.062\,\mathrm{m}\) for RL+PID, demonstrating better repeatability under hardware variability.
These gains arise from the hierarchical design: the RL outer loop generates task-level whole-body commands, while the estimator-based inner loop compensates for payload shifts, manipulator-induced coupling, and unmodeled dynamics. Overall, the results support the claim that learned whole-body coordination is more effective when paired with estimator-based low-level compensation, especially in high-speed and high-payload regimes. Since the comparison is limited to the figure-eight hardware study and does not include formal significance testing, the results should be interpreted as strong empirical evidence.

\section{Conclusion}
We presented a hierarchical control framework for aerial manipulation that combines an RL outer loop for task-level whole-body command generation with a delay-based inner-loop dynamics estimator for low-level uncertainty compensation. Hardware experiments demonstrate that the proposed method consistently outperforms RL+PID and RL+INDI+PID, achieving the lowest tracking error across all tested conditions, improving mean RMSE by up to \(40.9\%\), and yielding the smallest trial-to-trial variation. These results show that coupling learned whole-body coordination with estimator-based inner-loop compensation substantially improves precision, robustness, and repeatability under changing payload and motion conditions.

% The present study is limited to controlled hardware experiments and a baseline comparison centered on the figure-eight trajectory. Future work will extend the evaluation to broader trajectory families, more diverse manipulation tasks, and outdoor conditions with external disturbances. Nevertheless, the current results already indicate that combining RL with estimator-based low-level compensation is a promising direction for robust, high-performance aerial manipulation in uncertain environments.

% \section*{ACKNOWLEDGMENT}

% The authors would like to acknowledge the use of large language models (Gemini, ChatGPT) solely for refining and improving the linguistic flow of this manuscript.

\bibliographystyle{IEEEtran}
\bibliography{CASE}

\end{document}